\documentclass{article}

\usepackage{arxiv}

\usepackage{hyperref}
\usepackage{url}            
\usepackage{booktabs}       
\usepackage{amsfonts}       
\usepackage{nicefrac}       
\usepackage{microtype}      
\usepackage[T1]{fontenc}
\usepackage[english]{babel}
\usepackage[T1]{fontenc}
\usepackage{amsmath}
\usepackage{xcolor}
\usepackage{graphicx}
\usepackage{subcaption}

\definecolor{gpt_green}{RGB}{102,194,165}
\definecolor{claude_red}{RGB}{252,141,98}
\definecolor{ocr_purple}{RGB}{141,160,203}
\definecolor{ft_pink}{RGB}{231,138,195}
\definecolor{onlystripped_gray}{RGB}{0,0,0} 
\definecolor{lowered_green}{RGB}{143,206,0} 
\definecolor{unidecoded_blue}{RGB}{67, 162, 202} 
\definecolor{normalized_pink}{RGB}{250, 159, 181} 

\usepackage{color}

\newcommand{\edit}[1]{\textcolor{black}{#1}}
\newcommand{\editt}[1]{\textcolor{black}{#1}}

\title{Early evidence of how LLMs outperform traditional systems on OCR/HTR tasks for historical records}

\author{
 Seorin Kim \\
  Data Analytics Laboratory\\
  Vrije Universiteit Brussel (VUB)\\
  Brussels, Belgium \\
  \texttt{seorin.kim@vub.be} \\
   \And
 Julien Baudru \\
  IRIDIA, FARI\\
  Université Libre de Bruxelles (ULB)\\
  Brussels, Belgium \\
  \texttt{julien.baudru@ulb.be} \\
  \And
 Wouter Ryckbosch \\
  Social History of Capitalism\\
  Vrije Universiteit Brussel (VUB)\\
  Brussels, Belgium \\
  \texttt{} \\
    \And
 Hugues Bersini \\
  IRIDIA, FARI\\
  Université Libre de Bruxelles (ULB)\\
  Brussels, Belgium \\
  \texttt{} \\
  \And
 Vincent Ginis \\
  Data Analytics Laboratory\\
  Vrije Universiteit Brussel (VUB)\\
  Brussels, Belgium \\
  \texttt{} \\
}

\begin{document}
\maketitle
\begin{abstract}

 We explore the ability of two LLMs -- GPT-4o and Claude Sonnet 3.5 -- to transcribe historical handwritten documents in a tabular format and compare their performance to traditional OCR/HTR systems: EasyOCR, Keras, Pytesseract, and TrOCR. Considering the tabular form of the data, two types of experiments are executed: one where the images are split line by line and the other where the entire scan is used as input. Based on CER and BLEU, we demonstrate that LLMs outperform the conventional OCR/HTR methods. Moreover, we also compare the evaluated CER and BLEU scores to human evaluations to better judge the outputs of whole-scan experiments and understand influential factors for CER and BLEU. Combining judgments from all the evaluation metrics, we conclude that two-shot GPT-4o for line-by-line images and two-shot Claude Sonnet 3.5 for whole-scan images yield the transcriptions of the historical records most similar to the ground truth.
 

\end{abstract}

\section{Introduction}

Digitalizing historical documents has been a \edit{tedious} task for many governments and has allowed research opportunities for many. Besides manual efforts, optical character recognition (OCR) and handwritten text recognition (HTR) models have been the mainstays of digitizing historical documents. Given the increasingly expanding capabilities of the large language models (LLMs), we aim to study how much of the transcription process can be facilitated by the sole use of LLMs compared to that of the traditional OCR/HTR \edit{pipelines}. For this study, we focus on a tabulated historical record where the table's header is filled with optical characters while the rest contains handwritten texts. 

Historical records are more challenging to transcribe than modern documents due to cursive handwriting styles, \editt{the degraded quality of the texts (e.g., faded inks or damaged paper)}, language changes, and document layouts. The current OCR/HTR pipelines typically begin by pre-processing scans of the documents by adjusting the contrast and color of the image and cropping if necessary. Then, they analyze their layout and segment and then recognize the texts. In addition, layout analysis/segmentation and text recognition often accompany fine-tuning the user's data set. Unlike the three-step process of the classical OCR/HTR pipelines, LLMs can provide a one-step solution from the user's view, where only the final outputs need to be checked. Yet, their performability compared to the conventional methods \edit{remains unexplored}.

In this study, we aim \edit{to compare LLMs and OCR/HTR tools in terms of} how accurately they reproduce the texts in the scanned images of the historical records. Note that our goal is to replicate the texts in the image as they are, not to understand and guess what the information should be in the table. We utilize Character Error Rate (CER) and Bilingual Evaluation Understudy (BLEU) to evaluate the performance of the models: GPT-4o, Claude Sonnet 3.5, EasyOCR, Keras, Pytesseract, and TrOCR. The study also compares different strategies to enhance LLMs' transcription performance and \editt{two TrOCR variants with fine-tuning.}



\section{Similar work}

In \cite{dtrocr}, the authors introduce a new method for text recognition called Decoder-only Transformer for Optical Character Recognition (DTrOCR). DTrOCR uses only a decoder, using a pre-trained generative language model, in contrast to traditional encoder-decoder methods \edit{\cite{vaswani2023attentionneed}}. The authors tested whether a successful natural language processing model could be applied to text recognition in computer vision. Their experiments showed that DTrOCR significantly outperformed current state-of-the-art methods in recognizing printed, handwritten, and scene text in both English and Chinese.

In \cite{Loffler2023}, the author introduces a method to digitize over 100,000 historic plans from the Swiss Archive for Landscaping Architecture using AI models. The approach employs a three-model architecture: a layout model to identify text, an OCR model to extract words, and a named entity recognition (NER) model to label key information. K-means clustering groups text blocks for OCR processing. Various deep-learning models were evaluated, including German BERT for NER, and retrained on the NVIDIA DGX-2 system. The pipeline achieved an F1 score of 48\%, with the NER model scoring 86\% and the OCR model correctly extracting 54\% of words. 

In \cite{fadeeva2024}, the authors propose a novel tokenized representation of digital ink for online handwriting recognition, addressing the shortcomings of naive OCR with vision-language models (VLMs). Integrating stroke sequences and images, this approach achieves state-of-the-art quality on three public datasets. Their findings show that VLMs benefit from multimodal inputs, that images are crucial when text representations are too long, and that multiple handwriting tasks can be combined effectively. The method is compatible with both parameter-efficient tuning and fine-tuning, suggesting future exploration of various handwriting task combinations in large VLMs.

The following studies have adopted LLMs in their OCR/HTR process as an assistant to correct the outputs after OCR. In \cite{post-ocr}, the authors address the challenge of poor OCR quality in digitized historical documents, which is a barrier to humanities research. Traditional post-OCR correction methods use sequence-to-sequence models. Instead, the authors propose using generative language models with a prompt-based approach. They demonstrate significant improvements in OCR error correction by tuning Llama 2 with prompts and comparing it to a fine-tuned BART model on 19th-century British newspaper articles. Llama 2 achieves a \edit{55}\% reduction in the CER, outperforming BART's \edit{23}\% reduction. This approach shows promise for researchers in improving the accessibility of historical texts by an LLM. 

Similarly, in \cite{boros-etal-2024-post}, the authors conducted a comparative study of the ability of fourteen LLMs to correct transcriptions produced using OCR, HTR, and ASR. They then evaluate these corrections by comparing them with ground truths from each document. They conclude that, although GPT-4 appears to be the best model among those tested, all the models degrade rather than improve transcriptions. And that, on the whole, LLMs are better at detecting errors than at correcting them, as they are subject to overcorrection.

While these studies have explored different methods to improve OCR systems, to our knowledge, \editt{not many studies have used LLMs as a main transcription tool. \cite{shi2023exploringocrcapabilitiesgpt4vision} was an earlier adopter in exploring the capabilities of GTP-4 Vision in several OCR and HTR tasks for English and Chinese datasets, and \cite{ghiriti2024exploring} used German typed text datasets to explore the same model for OCR tasks. However, since the experiments were conducted at the early stage of GPT-4 Vision, the authors in \cite{shi2023exploringocrcapabilitiesgpt4vision} used the web-based dialogue interface for the experiments, hindering them from examining different strategies, including prompting, few shots, and fine-tuning. Also, \cite{ghiriti2024exploring} did not explore different strategies, and the experiment was limited to reading typed texts rather than handwritten texts.} 

Therefore, our contribution lies in exploring the pure performability of LLMs in transcribing historical records\editt{, and different strategies to improve their performance.} 



\section{\edit{Methodology}}

\subsection{Dataset}
We use Belgian probation data from 1921. The dataset includes 20 scanned pages of Déclaration de Succession from Nivelles, a French-speaking \edit{city} in Belgium. \editt{The main language of the record is French, but some names are of Dutch-speaking Flemish origin.} All the scans share the same layout and column names, comprising two lines of typed text. \editt{Note that each scan contains names, some abbreviations (e.g., \textit{bte} for Baptiste, \textit{Dbre} for December), place names, including some municipalities that no longer exist, and numbers.}

\subsection{Metrics}\label{sec:metrics}
In our analysis, we consider human transcriptions to be the ground truth (GT). First, the draft is made by one of the authors and then the names of the deceased and the declared and the locations of the death that appeared in the document are verified with the online search environment of the State Archives of Belgium, \href{https://agatha.arch.be/search/genealogie/}{Agatha}, and the online genealogy database, \href{https://nl.geneanet.org/genealogie/}{Geneanet}. This draft is again checked by another author in the same manner. Once all the documents were transcribed, we reviewed the transcriptions again to harmonize the styles and agree on some uncertainties. Despite these thorough checks by the authors, errors may still exist in the GT data. Nonetheless, given that these transcriptions are consistent in quality and verified multiple times, we believe they are representative data to be compared with the LLM and OCR/HTR outputs.

We employ the CER and BLEU scores to evaluate the performance of different methods and models compared to the GT.

\subsubsection{CER}
In the OCR community, the CER metric was popularized by \cite{rice1996ocr}, and the exact method used in this research has been implemented in \cite{huggingface_evaluate}. The CER measures the similarity between the predicted transcription and the GT by calculating the edit distance (Levenshtein distance) between the two strings. The edit distance is the minimum number of single-character edits (insertions, deletions, or substitutions) required to change the predicted transcription into the GT. The equation of the CER is \edit{given by}

\begin{equation} \label{eq:cer} 
\text{CER} = \frac{S + D + I}{N} \edit{,}
\end{equation}

\edit{w}here $S$ represents the number of substitutions, $D$ denotes the number of deletions, $I$ indicates the number of insertions, and $N$ refers to the total number of characters in the GT.

A lower CER indicates a better match between the predicted transcription and the ground truth. Thus, a CER of 0\% represents a perfect match. The perfect match, here, implies matching the position and number of characters in two transcriptions as well as the spaces. For instance, for a predicted text \texttt{[Arrêté le vingt-et-un novembre 1919]} and a GT text \texttt{[ Arrêté le vingt-et-un novembre 1919 ]}, even though they resemble perfectly, due to the white space in the start and end of the candidate text, the CER is $2/35 \approx 0.057$, not 0. For this reason, CER is particularly suited for evaluating text recognition models where even minor character errors can significantly impact the readability and accuracy of the transcription.

\subsubsection{BLEU}
The BLEU score, introduced in \cite{papineni2002bleu}, is a metric for evaluating the quality of machine-translated text by comparing it to a set of reference translations. BLEU scores range between \(0\) and \(1\), with \(1\) indicating perfect similarity to the reference. The BLEU score is \edit{given by}

\begin{equation} \label{eq:bleueq}
\text{BLEU} = \text{BP} \cdot \exp \left( \sum_{n=1}^{N} w_n \log p_n \right) \edit{,}
\end{equation}

\edit{w}here \( \text{BP} \) is the brevity penalty, penalizing the candidate translations shorter than the reference, \( p_n \) is the precision for modified n-grams of order \( n \), and \( w_n \) is the weight for each \( n \)-gram (commonly set to \( \frac{1}{N} \) for balanced n-gram contributions). The precision \( p_n \) is \edit{given by}

\begin{equation} \label{eq:precision}
p_n = \frac{\sum_{g \in \text{Candidate}} \min(C(g), R(g))}{\sum_{g \in \text{Candidate}} C(g)} \edit{,}
\end{equation}

\edit{w}here \( C(g) \) is the count of n-gram \( g \) in the candidate, and \( R(g) \) is the count of n-gram \( g \) in the reference. The \( \text{BP} \) is \edit{given by}

\begin{equation} \label{eq:penalty}
\text{BP} = \begin{cases} 
1 & \text{if } c > r \\
\exp \left(1 - \frac{r}{c}\right) & \text{if } c \leq r \edit{,}
\end{cases}
\end{equation}

where \( c \) represents the length of the candidate and \( r \) the length of the reference. 

As BLEU uses n-gram, the word's position in the sentence or the number of spaces does not affect the score like in the CER. In the same example of the candidate text, \texttt{[Arrêté le vingt-et-un novembre 1919]} with respect to the reference text, \texttt{[ Arrêté le vingt-et-un novembre 1919 ]}, would score 1, a perfect match, in BLEU, which was around 0.057 in CER. However, since the accuracy of the n-grams matters, BLEU scores are more sensitive to the capitalization of the word and the symbols and accents in the word. Therefore, for the candidate texts like \texttt{[arrêté Le vingt et un Novembre 1919]}, only one uni-gram matches the reference text -- that is, \texttt{[1919]}, giving $1/7$ as a uni-gram precision, $p_1$ and 0 for more than bi-gram precisions. If we take the default $N = 4$, then the precisions are [$1/7, 0, 0, 0$], returning $0$ BLEU score since $\log{0}$ is undefined. Usually, when using BLEU scores for evaluation translations, multiple reference texts are given. However, in our case, since only one GT exists, the scores are generally lower than when multiple references exist. 

Although BLEU scores are known to resemble the human evaluations of translations' quality, whether this will also be true for transcription tasks where only one reference text is given is less known. Therefore, in addition to evaluating the LLM and OCR outputs, we will also examine the possible factors influencing the BLEU and CER scores in Section~\ref{sec:result}.

\subsection{Experiments} \label{sec:experi}
In our experiment, two approaches are adopted to evaluate the performances of LLMs and OCR/HTR systems: (1)~\editt{using each of the 20 whole scans as an input} and (2)~splitting the images line by line and feeding each line as an input. As mentioned in Section~\ref{sec:metrics}, at the start of the experiment, we created the GT dataset, which served as our reference standard. This dataset was carefully compiled to reflect the exact text and layout of the original documents in Excel sheets and then converted to a text format when using them. When evaluating the line-by-line outputs by LLMs and OCR/HTR systems, each row is split \edit{using} Adobe Photoshop, and the two-level header is considered in one row rather than two separate rows. Here, the splitting can also be \editt{automated} with Python or ImageMagick.

Once the GT dataset is done, we let LLMs read the document entirely or line by line, using five distinct strategies: \textit{simple prompt, complex prompt, one-shot, two-shot,} and \textit{refine}. This research compared two \edit{state-of-the-art} LLM models: Claude Sonnet 3.5 (20240620) and GPT-4o \edit{(gpt-4o-2024-08-06)}. The former was chosen because other versions, such as Haiku, regularly returned data privacy error messages, preventing us from obtaining the desired output. The latter, from OpenAI, was selected because some literature like \cite{boros-etal-2024-post} has already shown its great performance in correcting the OCR/HTR outputs. Note that when the document is read entirely by LLMs, the image size for the Claude Sonnet 3.5 model had to be reduced by a factor of 3 due to restrictions imposed by Anthropic. However, since the initial images were in very high resolution, their quality was not affected.

As summarized in Table \ref{tab:llm_prompts}, five strategies used different prompts except for one-shot and two-shots, which share the same prompt structure. We compare the impact of descriptive prompting by using a \textit{simple prompt} asking LLMs to recreate the table they see in the image and a \textit{complex prompt} with detailed descriptions of the document, both being zero-shot. \textit{One-shot} and \textit{two-shot} approaches are included to further analyze the impact of example-based guidance, providing the model with one or two example images and transcriptions, respectively. Additionally, despite our goal of replicating the historical records as they are, given that some degree of reasoning (e.g., \texttt{actif-passif = restant}, the family name of the deceased and the declarant may be the same) can facilitate the task, we also employ the \textit{refine} technique on the outputs with the complex prompt, allowing the model to improve its outputs iteratively. Note that sometimes, the LLMs may return an error with a message such as \verb|[Sorry, I cannot read the image]|. In this case, we force them to output something with the additional \textit{anti-error prompt}, as shown in Table \ref{tab:llm_prompts}. If an error message still exists as an output, we rerun that particular document in the case of the whole-scan experiment and that particular row in the case of the line-by-line experiment.

Simultaneously, we process the same set of scanned documents -- as a whole and line-by-line -- with four different OCR/HTR methods in their baseline, pre-trained states without any fine-tuning to observe the pure performances of these systems: EasyOCR, Pytesseract, Keras, and TrOCR. Note that for EasyOCR, we set the language to French. Given the conventional HTR pipelines with these systems where layout analysis and fine-tuning are involved, we also test TrOCR with two fine-tuning variants: one fine-tuned on 20\% and the other on 50\% of the data with 6 epochs. For a comparison, we will run the whole-scan experiment, fine-tuning 20\% of the data. This experiment requires a longer computation time than the line-by-line experiment, approximately two executive days. Considering this long computation time and poor performance, as we will discuss in the results section, fine-tuning 50\% of the whole-scan data is not performed.

After processing the pages with each method, we standardize all outputs first by utilizing the LLM with the post-processing prompt in Table \ref{tab:llm_prompts} and then by explicitly removing extraneous elements such as separators, delimiters, and any other non-textual components (e.g., hyphens and white spaces) that can interfere with our analysis. Additionally, some anomalies in the outputs are manually treated to ensure a fair and accurate comparison to the GT data (see details in \edit{the supplementary material}). Lastly, before calculating the BLEU and CER scores, the characters in the outputs and the GT are all \edit{are put in lowercase (\textit{lowered}), and any whitespace and accents are removed (\textit{stripped} and \textit{unidecoded})}. Note that this \edit{pre-processing} slightly improves the BLEU and CER scores (see \editt{Appendix} Fig.~\ref{fig:preprocessing_whole} for the comparisons).

\section{Results} \label{sec:result}

Before assessing the results, it should be noted that for the line-by-line experiments, we used the maximum order of n-grams of 3, while it is 4 for the whole-scan experiments. This decision was made since text lengths \editt{in the line-by-line experiments} are usually below 30 except for the header, which had 152 words. Having such a short prediction text, the maximum order of n-grams of 4 results in too many zero BLEU scores. Conversely, setting the maximum order of n-grams too low results in disproportionately high BLEU scores, complicating meaningful comparisons \editt{between} outputs. A comparison between bi-grams, tri-grams and 4-grams is presented in Appendix Fig.~\ref{fig:ngrams_perline}. 

\vspace{-1em}
\begin{figure*}[!thb]
\centering
\includegraphics[width=\textwidth]{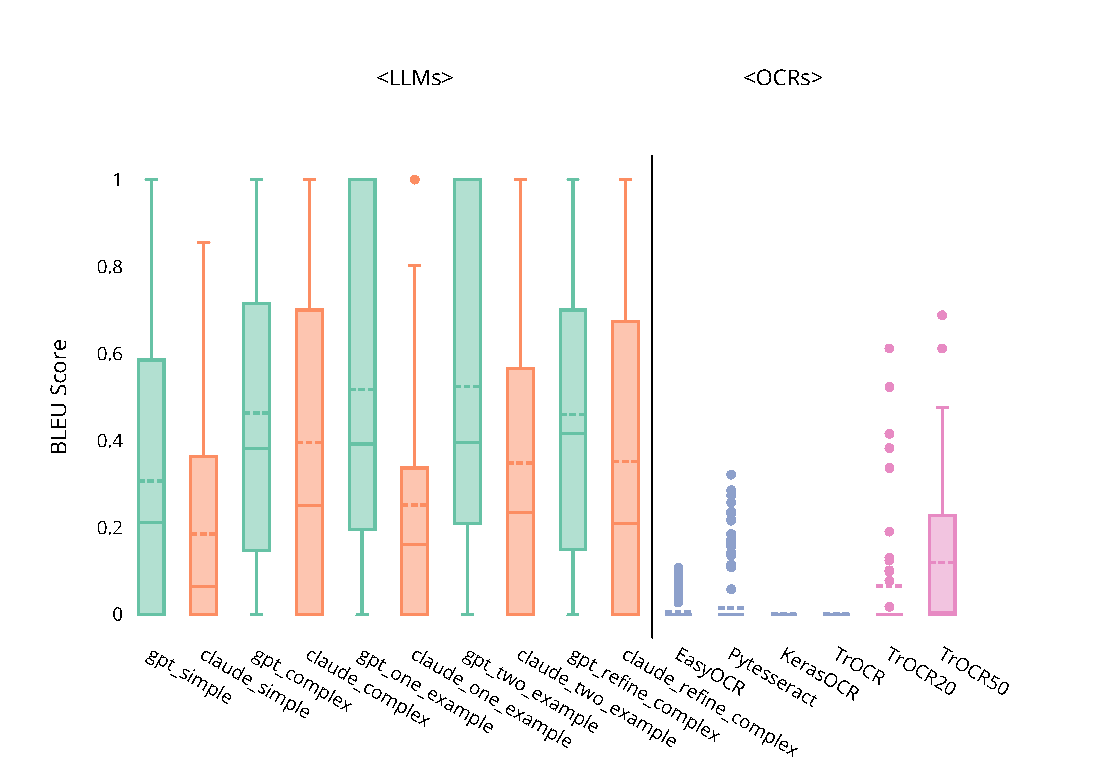}
\caption{BLEU score comparisons for each method on the line-by-line dataset: \raisebox{0.5ex}{\fcolorbox{gpt_green}{gpt_green!50}{\rule{0pt}{1pt}\rule{1pt}{0pt}}}~GPT 4-o, \raisebox{0.5ex}{\fcolorbox{claude_red}{claude_red!50}{\rule{0pt}{1pt}\rule{1pt}{0pt}}}~Claude Sonnet 3.5, \raisebox{0.5ex}{\fcolorbox{ocr_purple}{ocr_purple!50}{\rule{0pt}{1pt}\rule{1pt}{0pt}}}~OCR tools and \raisebox{0.5ex}{\fcolorbox{ft_pink}{ft_pink!50}{\rule{0pt}{1pt}\rule{1pt}{0pt}}}~Fine-tuned TrOCR. A higher BLEU means a higher similarity with the GT.
For each method, the mean score is in a dashed line, and the median is in a bold line. The maximum order in the n-gram used to calculate BLEU is 3. The texts are \edit{stripped, lowered and unidecoded} before calculating the scores.}
\label{fig:bleu_perline}
\end{figure*}

\begin{figure*}[!thb]
\centering
\includegraphics[width=\linewidth]{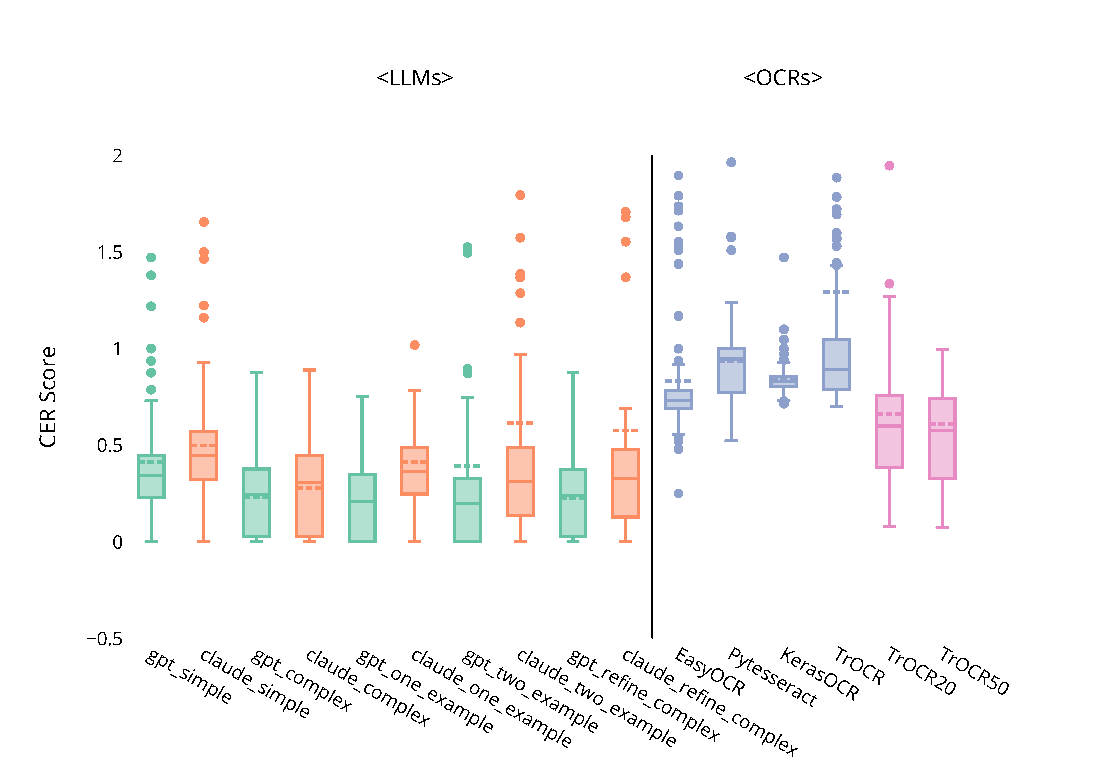}
\caption{CER score comparisons for each method on the line-by-line dataset: \raisebox{0.5ex}{\fcolorbox{gpt_green}{gpt_green!50}{\rule{0pt}{1pt}\rule{1pt}{0pt}}}~GPT 4-o, \raisebox{0.5ex}{\fcolorbox{claude_red}{claude_red!50}{\rule{0pt}{1pt}\rule{1pt}{0pt}}}~Claude Sonnet 3.5, \raisebox{0.5ex}{\fcolorbox{ocr_purple}{ocr_purple!50}{\rule{0pt}{1pt}\rule{1pt}{0pt}}}~OCR tools and \raisebox{0.5ex}{\fcolorbox{ft_pink}{ft_pink!50}{\rule{0pt}{1pt}\rule{1pt}{0pt}}}~Fine-tuned TrOCR. The Y-axis is zoomed at [-0.5, 2].  A lower CER means a higher similarity with the GT. The maximum CER value observed is 64. For each method, the mean score is in a dashed line, and the median is in a bold line. The texts are pre-processed as in Fig.~\ref{fig:bleu_perline}.}
\label{fig:cer_perline} 
\end{figure*}

When assessing the outputs of the LLMs and OCR/HTR systems, we observe that LLMs \edit{greatly} outperform the classical OCR/HTR methods in numerous ways. Firstly, comparing zero-shot LLMs and the OCR/HTR methods without fine-tuning, zero-shot LLMs could recreate the historical records with correctly read names, while the OCR/HTR methods without fine-tuning could not return any meaningful outputs. EasyOCR and Pytesseract could only read some words from the header, which is typed but not handwritten texts in both experimental types. KerasOCR's outputs comprise random orders of alphabets, and TrOCR outputs random English phrases in both types of experiments. 

Secondly, zero-shot LLMs still performed better than the fine-tuned TrOCR. Given that the OCR/HTR tools are conventionally used in the combination of layout analysis and fine-tuning, we expected the line-by-line experiments with 20\% and 50\% of the data trained (hereafter TrOCR20 and TrOCR50) to return outputs with high accuracy. Although the outputs of TrOCR20 and TrOCR50 resembled more of the GT with proper French words, the transcribed words often repeated the trained data, not properly transcribing the texts in the input image. 

Figs.~\ref{fig:bleu_perline} and \ref{fig:cer_perline} manifest the differences between the LLMs and OCR/HTR tools in BLEU and CER scores with the line-by-line experiments. For the BLEU score, the four OCR/HTR tools score very low, around 0, denoting no similarities between the GT and their outputs, while the fine-tuned TrOCR20 and TrOCR50, especially the latter, have higher scores than those without fine-tuning. Nevertheless, their mean scores are still lower than those of the LLM methods. Considering the CER metric, the differences between the LLMs and OCR/HTR tools are less marked than with the BLEU metric, albeit, on average, the LLMs have lower error values than the OCR/HTR systems.

\subsection{Whole-scan vs. Line-by-line experiments} 

\begin{figure*}[!htb]
\centering
\includegraphics[width=\linewidth]{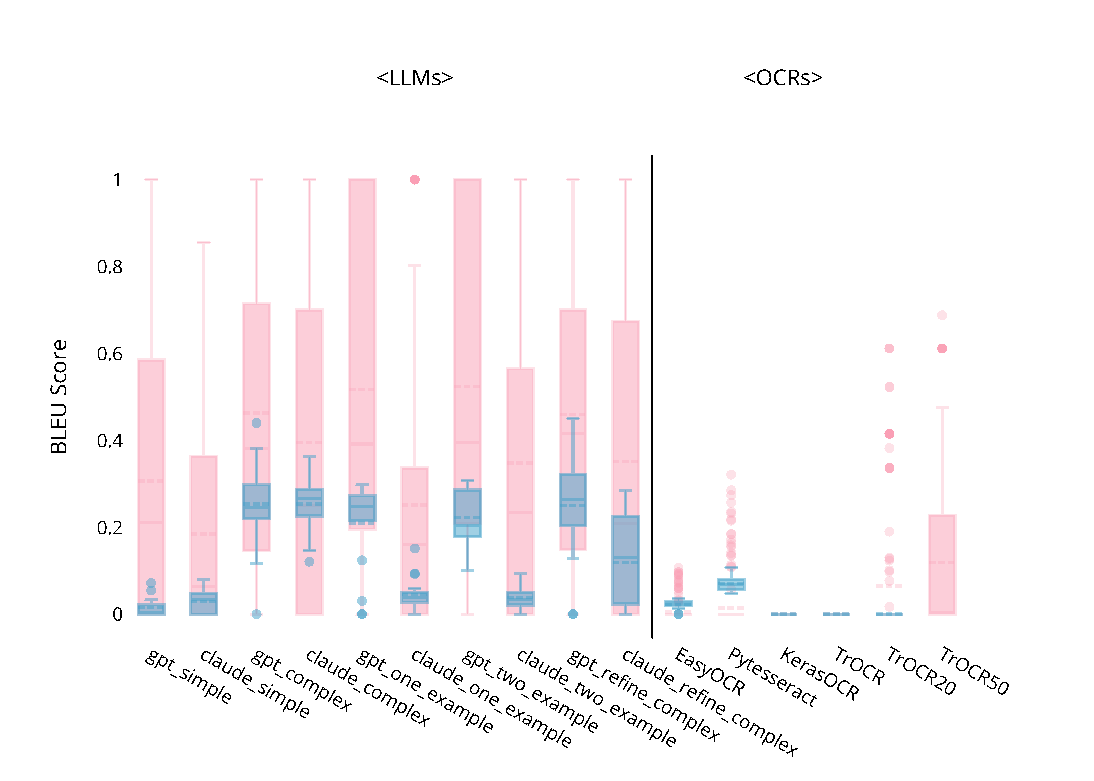}
\caption{\label{fig:bleu_whole} \edit{BLEU score comparisons of each method between the whole scan experiments (\raisebox{0.5ex}{\fcolorbox{unidecoded_blue}{unidecoded_blue!30}{\rule{0pt}{1pt}\rule{1pt}{0pt}}}), and the line-by-line experiments (\raisebox{0.5ex}{\fcolorbox{normalized_pink}{normalized_pink!30}{\rule{0pt}{1pt}\rule{1pt}{0pt}}}). A higher BLEU means a higher similarity with the GT. For each method, the mean score is in a dashed line, and the median is in a bold line. The maximum order in the n-gram used to calculate BLEU is 3 for the line-by-line experiments and 4 for the whole scan experiments. TrOCR50 is not performed with the whole scan dataset.}}
\end{figure*}

\begin{figure*}[!htb]
\centering
\includegraphics[width=\linewidth]{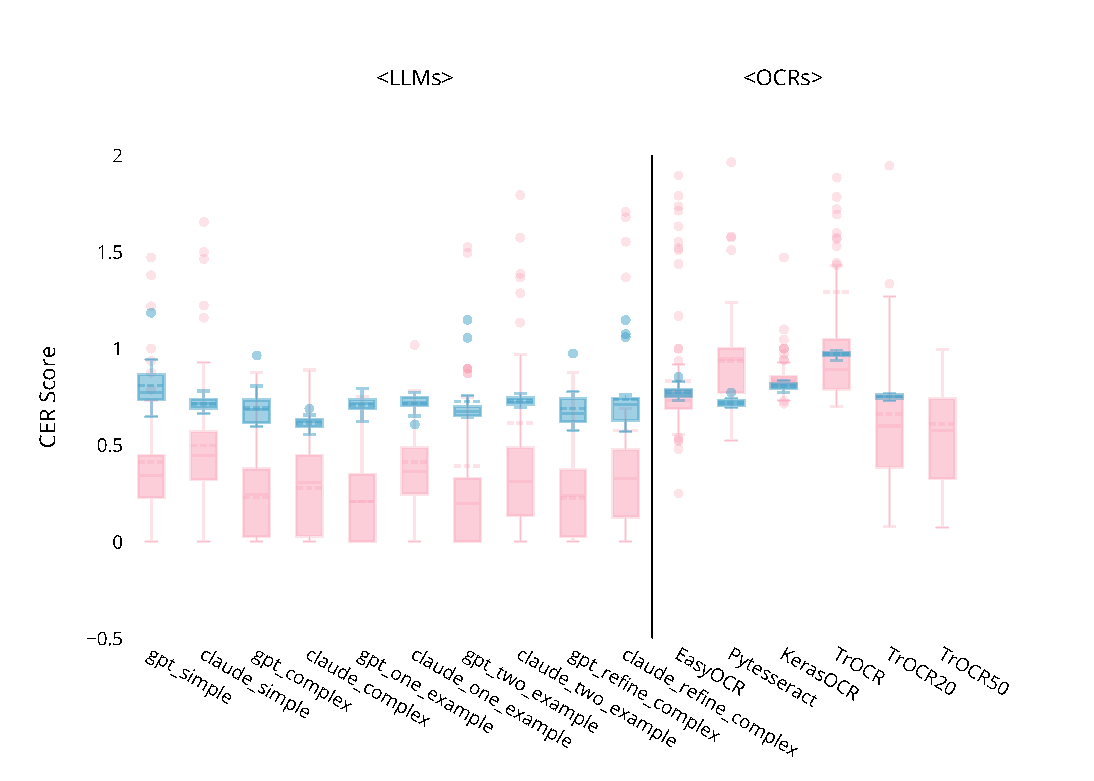}
\caption{\edit{CER score comparisons of each method between the whole scan experiments (\raisebox{0.5ex}{\fcolorbox{unidecoded_blue}{unidecoded_blue!30}{\rule{0pt}{1pt}\rule{1pt}{0pt}}}), and the line-by-line experiments (\raisebox{0.5ex}{\fcolorbox{normalized_pink}{normalized_pink!30}{\rule{0pt}{1pt}\rule{1pt}{0pt}}}). A lower CER means a higher similarity with the GT. The maximum CER value observed is 1.21 for the whole scan and 64 for the line-by-line experiments. The Y-axis is zoomed at [-0.5, 2]. For each method, the mean score is in a dashed line, and the median is in a bold line. TrOCR50 is not performed with the whole scan dataset.}}
\label{fig:cer_whole}
\end{figure*}

When comparing BLEU scores between the two types of experiments, we notice the smaller variances in BLEU scores in the whole-scan experiments \edit{than in the line-by-line experiments in Fig.~\ref{fig:bleu_whole}}. This may be due to the fact that the sample size is larger in the line-by-line experiments than in the whole-scan experiments (283 vs. 20). Moreover, the header, which is typed and remains the same over 20 documents, is often well transcribed in the line-by-line experiments. Therefore, if each of the 20 headers is perfectly transcribed, it increases the maximum BLEU score of the method and decreases its minimum CER score. 

Despite the larger variances, the line-by-line experiments score higher on average than the whole-scan experiments. The two OCR/HTR tools without any fine-tuning -- EasyOCR and Pytesseract -- which scored nearly zero in the line-by-line experiments, scored slightly better \editt{in the whole-scan experiments} than TrOCR20. Yet, the outputs do not look as convincing as those by LLMs. They did not manage to read the handwritten texts and mainly read the typed texts in the table's header, albeit not perfectly. This may hint that without the fine-tuning and cropping of the images, EasyOCR and Pytesseract perform better than KerasOCR and TrOCR, but their performance is limited to OCR rather than HTR. 
\editt{Moreover, contrary to the traditional strategy of applying a layout analysis and segmentation to improve the performance, we did not see a significant improvement when using these OCR/HTR tools with sliced words of one scanned document. Besides the lower performance, such an approach required more pre-processing, resulting in lower efficiency.}

\editt{Regarding CER scores, while the medians and means of the CER scores are generally lower in line-by-line experiments, denoting higher similarities with the GT, the whole-scan experiments demonstrate smaller variances than the line-by-line experiments in Fig.~\ref{fig:cer_whole}. Moreover, compared to the BLEU scores, the CER scores do not differ much across the methods in the whole-scan experiments. This is because of two reasons. First, the whole-scan document's GT contains more text than when separating them into each row, which results in a larger denominator in Eq.~\ref{eq:cer}. Second, feeding the whole scan as an input does not always guarantee the correct number of rows in the results. Mostly, the outputs, regardless of the methods, returned fewer characters than the GT, which penalizes all the methods similarly in the CER calculation. However, unlike the similar CER scores across all the methods, we noticed that LLM outputs contain more words similar to those in GT than OCR outputs. In the next section, we will delve deeper into the factors that cause this phenomenon.}

\subsection{BLEU vs. CER vs. Human Evaluations}
As observed in the previous subsections, BLEU and CER scores do not always concord. Especially when comparing LLMs to OCRs, BLEU scores in both experimental types show more distinct differences between LLMs and OCRs than CER scores. This difference arises from their different approaches to measuring similarities between the reference and candidate text, as \edit{discussed} in Section~\ref{sec:metrics}. The CER score measures the similarities by counting each correct character, including spaces, and how many insertions and deletions it requires to achieve the correct text. The BLEU score, however, focuses on n-grams of words, regardless of their positions in the text. Accordingly, the CER is sensitive to the white spaces, and the BLEU is sensitive to the normalization of the words and the choice of the maximum order in n-grams. 

Given the weakness of the two scores, \edit{we organized evaluations} for whole-scan experiments to investigate which metric better represents the transcription quality according to human \edit{judgment}. When \edit{the two lead authors} evaluated the outputs of fourteen methods from whole-scan experiments, the two-example prompt by Claude Sonnet 3.5 returned the best outputs in terms of format and content, the detailed results of these evaluations are given in Table \ref{tab:human_evaluation}. Note that according to BLEU, it was GPT-4o with the \textit{refine} prompt, and it was Claude Sonnet 3.5 with the \textit{complex} prompt according to CER. 

The differences arise from the way human evaluators weigh certain information more than others. First, human evaluators are lenient with the outputs where the header is not at all or not well transcribed. Human evaluators focus more on the handwritten content, which is arguably more important than the header to transcribe, given the end goal of using the transcriptions in an analysis. This is advantageous for Claude Sonnet 3.5. who often returned outputs without having transcribed the header. However, such outputs without the header lose lots of scores in BLEU and increase CER since the header takes up 152 correct words and 828 correct characters. 

Second, \edit{the} human evaluators \edit{were} also more lenient with wrong numbers and \edit{weighed} more on correctly transcribing letters such as names of the deceased, locations, and dates. Since the evaluators created the GT dataset, they were more understanding of the poor performance in transcribing the numbers \edit{given their low readability}. Therefore, while each wrong number lowers a BLEU score and the wrong digit increases a CER, their influences are mere on the human evaluation. 

\begin{table}[!thb]
\centering
\caption{\label{tab:human_evaluation}Average scores of 20 scans by human evaluators. The score ranges from 1 being Very Bad and 5 being Very Good.}
\begin{tabular}{lc}
\hline
\textbf{Method}       & \textbf{Average Score} \\ \hline
GPT simple            & 1.08                   \\
Claude simple         & 2.13                   \\
GPT complex           & 1.33                   \\
Claude complex        & 2.28                   \\
GPT one example         & 1.66                   \\
\textbf{Claude one example}      & \textbf{3.55}            \\
GPT two example        & 1.53                   \\
\textbf{Claude two example}     & \textbf{4.06}                  \\
GPT refine complex    & 1.40                   \\
Claude refine complex & 2.88                   \\
KerasOCR              & 1.00                   \\
Pytesseract OCR       & 1.00                   \\
EasyOCR               & 1.00                   \\
TrOCR                 & 1.00                   \\ \hline
\end{tabular}
\end{table}

In order to examine the impact of the headers on the whole-scan experiments' BLEU and CER scores, we manually disregard the headers in the whole-scan outputs to compare them to the GT without the headers. Since this step involves personal judgment, we only worked with the best two outputs according to human evaluation (i.e., Claude two-example and Claude one-example) and their counterparts (i.e., GPT two-example and GPT one-example). 

\begin{figure}[!th]
\begin{subfigure}[b]{0.49\linewidth}
    \includegraphics[width=1.05\textwidth]{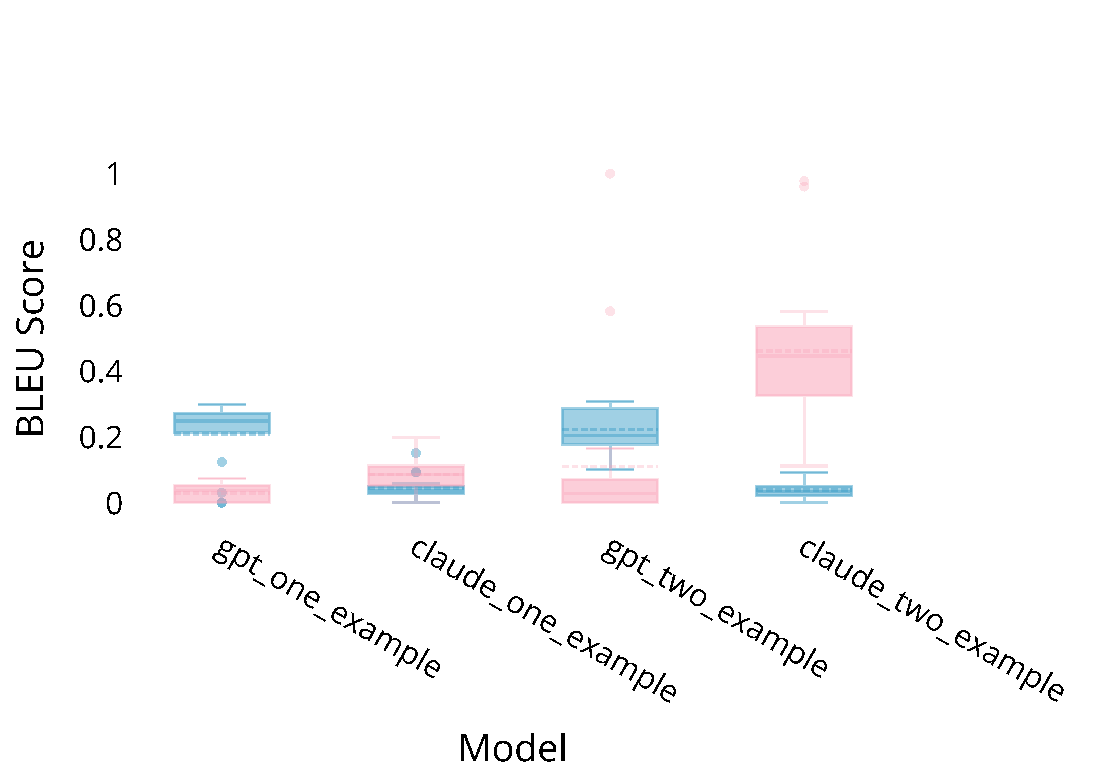}
    \caption{}
\end{subfigure}
\hfill
\begin{subfigure}[b]{0.49\linewidth}
    \includegraphics[width=1.05\textwidth]{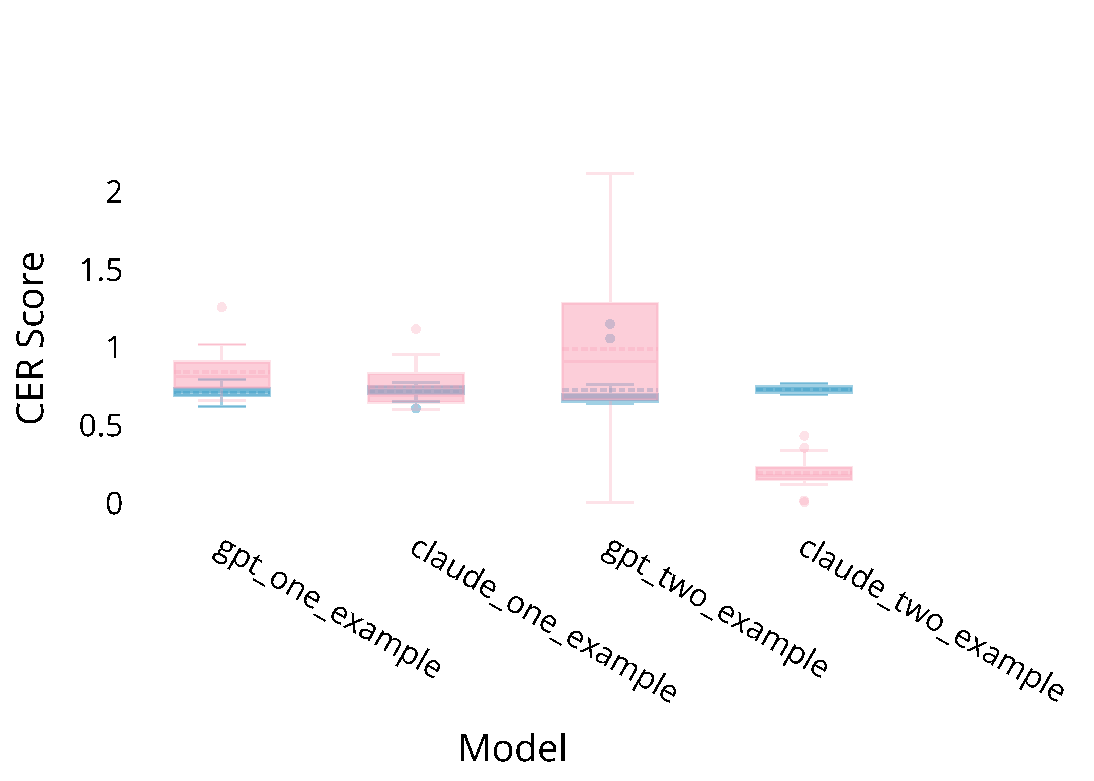}
    \caption{}
\end{subfigure}
\caption{BLEU and CER scores with the headers (\raisebox{0.5ex}{\fcolorbox{unidecoded_blue}{unidecoded_blue!30}{\rule{0pt}{1pt}\rule{1pt}{0pt}}})  and without the headers (\raisebox{0.5ex}{\fcolorbox{normalized_pink}{normalized_pink!30}{\rule{0pt}{1pt}\rule{1pt}{0pt}}}) for one-example and two-example prompts by GPT-4o and Claude Sonnet 3.5. After removing the headers, Claude Two-Example has the highest BLEU and lowest CER on average.}
\label{fig:whole_noheader}
\end{figure}

Fig.~\ref{fig:whole_noheader} shows that by disregarding the headers, BLEU scores of the two methods with Claude increase -- Claude two-example in particular -- while GPT methods' BLEU scores decreased. Similarly, the CER scores of the Claude two-example case dramatically reduced from a median of 0.72 to 0.17. In comparison, the effect is not as big in the Claude one-example case, where the median CER score still reduced from 0.72 to 0.69 but with a larger variance than when the headers were included. For the two GPT cases, both of their CER scores increased when the headers were disregarded in the evaluation, and their BLEU scores decreased. In the line-by-line experiments, the impact was not as big as in the whole-scan experiments. Still, GPT-4o with the two-example prompt showed the largest BLEU.

Having seen the large fluctuations in the scores before and after disregarding the headers, we find understanding the structure of the outputs is crucial if one aims to compare the quality of transcriptions with more weight on particular information. In that case, one can also modify BLEU and CER further according to their needs. Moreover, between BLEU and CER metrics, BLEU seems to distinguish the quality of the transcriptions better than CER.

\section{Discussion and Future Work}\label{sec:discussion}

In this study, we only explored the \edit{historical} transcription performability of two LLMS: GPT-4o and Claude Sonnet 3.5. As new LLMs are developing at a high speed, we \edit{forsee} that LLMs' performability in \edit{historical} transcription will increase. Moreover, this study \edit{focused on examining the performance of a single model. However, g}iven that each LLM has \edit{unique} strong points, one can also combine different \edit{LLMs} to examine whether such an approach enhances the performance and to what extent. For example, employing GPT-4 for preliminary image text recognition and subsequently using GPT-3.5 for refining or error correction could offer a synergistic approach to improving overall performance, in a similar way to the method proposed in \cite{xu2024llmrefinepinpointingrefininglarge}.

While conducting the experiments over several months, we found that the LLMs, especially GPT-4o, changed their behavior several times. For instance, an apologetic message that we had never received during the pilot experiment appeared more frequently at the end of the experiment. Although we had solutions to this issue by including an extra prompt and rerunning if necessary, we recognize that this hinders the reproduction of the exact outputs. \edit{Therefore, future researchers should consider the models and prompts detailed in the Appendix.}

Another challenge that we faced throughout the study is that LLMs have a hard time reading digits. Especially when one aims to utilize digitized records for quantitative research, transcribing the digits correctly is crucial. Considering that multi-modal fine-tuning becomes increasingly possible, future work may consider fine-tuning for digits \editt{or different strategies to overcome this issue}.

Similarly, this study only considered the sole goal of accurately reading the texts from images rather than correctly interpreting the information in the images. While this goal allowed us to focus on the pure performability of LLMs, future work can delve further into the other goal, which aims to increase the data quality by detecting and correcting misspellings or omission of information. 

In this context, one can consider adopting different output formats like JSON or XML. According to \cite{tam2024let}, format constraints like JSON or XML significantly impair LLM's reasoning abilities, while the restricted formats increase their performance in classification tasks. \edit{Focusing on the data quality, the task involves not only the classification task, classifying each image of a text with a text, but also the reasoning task, detecting the rules in the document and reasoning about possible errors in the read text. Therefore, it will be interesting to compare which approach retains outputs with higher data quality.}


Another potential avenue for future improvement, as briefly discussed in Section~\ref{sec:experi}, involves the fine-tuning of the TrOCR model. Specifically, it would be valuable to evaluate the performance of the LLM-based strategies against a version of TrOCR trained on a larger dataset and subjected to additional training epochs. However, achieving this enhancement would necessitate the creation of a more extensive set of GT data, a process that is both time-intensive and potentially costly due to its reliance on manual effort.

Finally, future research should focus on comparing the OCR/HTR systems and the LLM-based strategies used in this study against well-established benchmarks, such as the IAM dataset \cite{marti2002iam} and the READ16 dataset \cite{toselli2018htr}, to further assess their relative performance.

\section{Conclusion}
Our experiments have demonstrated the potential of LLMs to transcribe historical hand-written documents efficiently. \editt{Notably, the LLMs seem to perform better when the document images are sliced per row rather} than when whole scans are used as inputs. Among the two LLMs tested, Claude Sonnet 3.5 performed better with the whole scans, and GPT-4o performed better with the sliced images. Regarding the strategies, giving two examples with a prompt yielded the best results in both cases.

\editt{For historians, an important aspect of the OCR/HTR tools has been that originally, there was very little ground truth data available, and thus, very few models could be applied broadly, adopting different handwriting styles, languages, and source types. In our experiments, however, LLMs demonstrated the ability to produce highly accurate outputs with as few as two pieces of ground truth data. Furthermore, neither tabulated data formats nor variations in handwriting styles within the same page posed significant obstacles for these models.}


When evaluating the similarities between the ground truth and the outputs, it appeared crucial to understand the characteristics of the documents. For transcription tasks, as in our experiments, we observed that the BLEU score captures the distinctiveness in output qualities between different methods better than the CER. \editt{This highlights the potential of the BLEU score as a metric for evaluating OCR/HTR tasks involving longer texts. Moreover, since the document header was repetitive and less critical to transcribe accurately than the content, disregarding the headers during evaluation allowed for a more meaningful ranking of output quality across methods.}

To conclude, the study demonstrates that the LLMs outperform the conventional OCR/HTR tools in historical transcription based on the state-of-the-art metrics, BLEU, and CER scores. Notably, this superior performance is observed even though these metrics tend to underestimate the capabilities of LLMs.

\section*{Acknowledgement}
This publication is supported by the Communauté française de Belgique as part of an FRIA grant, by the FARI - AI for the Common Good Institute (ULB-VUB), by the European Union, with the support of the Brussels Capital Region (Innoviris and Paradigm), and by the Flemish interuniversity iBOF programme (An Ancestor’s tale).



%
%
%
\bibliographystyle{splncs04}
\bibliography{main}

\begin{thebibliography}{10}
\providecommand{\url}[1]{\texttt{#1}}
\providecommand{\urlprefix}{URL }
\providecommand{\doi}[1]{https://doi.org/#1}

\bibitem{boros-etal-2024-post}
Boros, E., Ehrmann, M., Romanello, M., Najem-Meyer, S., Kaplan, F.: Post-correction of historical text transcripts with large language models: An exploratory study. In: Bizzoni, Y., Degaetano-Ortlieb, S., Kazantseva, A., Szpakowicz, S. (eds.) Proceedings of the 8th Joint SIGHUM Workshop on Computational Linguistics for Cultural Heritage, Social Sciences, Humanities and Literature (LaTeCH-CLfL 2024). pp. 133--159. Association for Computational Linguistics, St. Julians, Malta (Mar 2024), \url{https://aclanthology.org/2024.latechclfl-1.14}

\bibitem{fadeeva2024}
Fadeeva, A., Schlattner, P., Maksai, A., Collier, M., Kokiopoulou, E., Berent, J., Musat, C.: Representing online handwriting for recognition in large vision-language models (2024), \url{https://arxiv.org/abs/2402.15307}

\bibitem{dtrocr}
Fujitake, M.: Dtrocr: Decoder-only transformer for optical character recognition (2023)

\bibitem{ghiriti2024exploring}
Ghiriti, A., G{\"o}derle, W., Kern, R.: Exploring the capabilities of gpt4-vision as ocr engine. In: International Conference on Theory and Practice of Digital Libraries. pp. 3--12. Springer (2024)

\bibitem{huggingface_evaluate}
HuggingFace: Evaluate: A library for easily evaluating machine learning models and datasets. \url{https://github.com/huggingface/evaluate} (2024), \url{https://github.com/huggingface/evaluate}, version 0.4.3, Apache-2.0 License

\bibitem{Loffler2023}
Löffler, K.: Digitize Historic Architectural Plans with OCR and NER Transformer Models. Other thesis, OST Ostschweizer Fachhochschule (May 2023), \url{https://eprints.ost.ch/id/eprint/1189}, thesis advisor: Mitra Purandare

\bibitem{marti2002iam}
Marti, U.V., Bunke, H.: The iam-database: an english sentence database for offline handwriting recognition. International Journal on Document Analysis and Recognition  \textbf{5}(1),  39--46 (2002)

\bibitem{papineni2002bleu}
Papineni, K., Roukos, S., Ward, T., Zhu, W.J.: Bleu: a method for automatic evaluation of machine translation. In: Proceedings of the 40th annual meeting of the Association for Computational Linguistics. pp. 311--318 (2002)

\bibitem{rice1996ocr}
Rice, S.V., Jenkins, F.R., Nartker, T.A.: The fifth annual test of ocr accuracy. Tech. rep., University of Nevada, Las Vegas (1996), \url{https://www.psu.edu}

\bibitem{shi2023exploringocrcapabilitiesgpt4vision}
Shi, Y., Peng, D., Liao, W., Lin, Z., Chen, X., Liu, C., Zhang, Y., Jin, L.: Exploring ocr capabilities of gpt-4v(ision) : A quantitative and in-depth evaluation (2023), \url{https://arxiv.org/abs/2310.16809}

\bibitem{tam2024let}
Tam, Z.R., Wu, C.K., Tsai, Y.L., Lin, C.Y., yi~Lee, H., Chen, Y.N.: Let me speak freely? a study on the impact of format restrictions on performance of large language models (2024), equal contribution, Equal advisorship

\bibitem{post-ocr}
Thomas, A., Gaizauskas, R., Lu, H.: Leveraging {LLM}s for post-{OCR} correction of historical newspapers. In: Sprugnoli, R., Passarotti, M. (eds.) Proceedings of the Third Workshop on Language Technologies for Historical and Ancient Languages (LT4HALA) @ LREC-COLING-2024. pp. 116--121. ELRA and ICCL, Torino, Italia (May 2024), \url{https://aclanthology.org/2024.lt4hala-1.14}

\bibitem{toselli2018htr}
Toselli, A.H., Romero, V., Villegas, M., Vidal, E., Sánchez, J.A.: Htr dataset icfhr 2016 (version 1.2.0) (2018). \doi{10.5281/zenodo.1297399}, \url{https://zenodo.org/record/218236}, dataset from the READ project (Horizon 2020) containing annotated pages from the Ratsprotokolle collection in Early Modern German.

\bibitem{vaswani2023attentionneed}
Vaswani, A., Shazeer, N., Parmar, N., Uszkoreit, J., Jones, L., Gomez, A.N., Kaiser, L., Polosukhin, I.: Attention is all you need (2023), \url{https://arxiv.org/abs/1706.03762}

\bibitem{xu2024llmrefinepinpointingrefininglarge}
Xu, W., Deutsch, D., Finkelstein, M., Juraska, J., Zhang, B., Liu, Z., Wang, W.Y., Li, L., Freitag, M.: Llmrefine: Pinpointing and refining large language models via fine-grained actionable feedback (2024), \url{https://arxiv.org/abs/2311.09336}

\end{thebibliography}

\newpage
\appendix
\section{Appendix} \label{sec:appendix}

\begin{table}[!ht]
    \centering
    \caption{\edit{Prompts for Claude Sonnet 3.5 (20240620) and GPT-4o-2024-08-06}}
    \resizebox{0.8\textwidth}{!}{%
    \begin{tabular}{c|ll}
        \cline{1-2}
        \textbf{Strategy} & \multicolumn{1}{c}{\textbf{Prompt}}       &  \\ \hline
        Simple prompt     & \texttt{Recreate the table you see in the image.} &  \\ \hline
        Complex prompt    & \begin{tabular}[c]{@{}l@{}}%
        \texttt{From the example, you learned the handwriting of this Belgian record.} \\ 
        \texttt{You learned which alphabet and which number is written in which way.}\\ 
        \texttt{With this knowledge, now consider the following image to recreate:}\\ 
        \texttt{First, you read a two-level header in the table, which you recognize} \\ 
        \texttt{the same as the example as follows in the form}\\ 
        \texttt{of ("first level", "second level"):}\\ 
        \texttt{```}\\ 
        \texttt{[("N' d'ordre", " "),}\\ 
        \texttt{("Date du dépot des déclarations", " "),}\\ 
        \texttt{("Désignation des personnes décédées ou absentes.:", "Nom."),}\\ 
        \texttt{("Désignation des personnes décédées ou absentes.:", "Prénoms"),}\\ 
        \texttt{("Désignation des personnes décédées ou absentes.:", "Domiciles"),}\\ 
        \texttt{("Date du décès ou du jugement d'envoi en possession, en cas d'absence.", " "),} \\ 
        \texttt{("Noms, Prénoms et demeures des parties déclarantes.", " "),}\\ 
        \texttt{("Droits de succession en ligne collatérale et de mutation en ligne directe.", } \\ 
        \texttt{"Actif. (2)"),}\\ 
        \texttt{("Droits de succession en ligne collatérale et de mutation en ligne directe.", } \\ 
        \texttt{"Passif. (2)"),}\\ 
        \texttt{("Droits de succession en ligne collatérale et de mutation en ligne directe.", } \\ 
        \texttt{"Restant NET. (2)"),}\\ 
        \texttt{("Droit de mutation par déces", "Valeur des immeubles. (2)"),}\\ 
        \texttt{("Numéros des déclarations", "Primitives."),}\\ 
        \texttt{("Numéros des déclarations", "Supplémentaires."),}\\ 
        \texttt{("Date", "de l'expiration du délai de rectification."),}\\ 
        \texttt{("Date", "de l'exigibilité des droits."),}\\ 
        \texttt{("Numéros de la consignation des droits au sommier n' 28", " "),}\\ 
        \texttt{("Recette des droits et amendes.", "Date"),}\\ 
        \texttt{("Recette des droits et amendes.", "N\textasciicircum{}03"),}\\ 
        \texttt{("Cautionnements.", "Numéros de la consignation au sommier n'30"),}\\ 
        \texttt{("Observations (les déclarations qui figurent à l'état n'413 doivent} \\ 
        \texttt{être émargées en conséquence, dans la présente colonne.)", " ")]}\\ 
        \texttt{```}\\ 
        \texttt{Context:}\\ 
        \texttt{- It's written in French language and the names of the people are} \\ 
        \texttt{domiciles are Belgian.}\\ 
        \texttt{- Each row contains information about a dead person for the 20} \\ 
        \texttt{variables above.}\\ 
        \texttt{Some rows contain information about the service date of the} \\ 
        \texttt{dead person written in the previous row.}\\ 
        \texttt{Such rows begin with texts like "Arrêté le \textbackslash{}d\{2\} \textbackslash{}w+ \textbackslash{}d\{4\}(\textbackslash{}w+)? servais"} \\ 
        \texttt{under "Nom." variable.}\\ 
        \texttt{- When you see "Arrêté le \textbackslash{}d\{2\} \textbackslash{}w+ \textbackslash{}d\{4\}( \textbackslash{}w+)? servais", the}\\ 
        \texttt{subsequent row will be the next serviced day.}\\ 
        \texttt{- N' d'ordre will also follow an order.}\\ 
        \texttt{- The family name in this column "Noms, Prénoms et demeures} \\ 
        \texttt{des parties déclarantes." may be the same as the family name in "Nom." column.}\\ 
        \texttt{Task:}\\ 
        \texttt{Please recreate the table by filling in all the information in the record.}\\ 
        \texttt{Pay attention to reading each word and number correctly.}\\ 
        \texttt{```plaintext}\\ 
        \texttt{"""}%
        \end{tabular} &  \\ \hline
        One-/Two-shots  & \begin{tabular}[c]{@{}l@{}}%
        \texttt{The ```plaintext block is the example transcription of the example image you saw:}\\ 
        \texttt{Transcription:}\\ 
        \texttt{```plaintext}\\ 
        \texttt{\{example(s)\}}\\ 
        \texttt{```}\\ 
        \texttt{Compare what you read initially and the solution key in ```plaintext block.}\\ 
        \texttt{Recreate the content of the table in this image. Only that,} \\
        \texttt{no other information from you.}%
        \end{tabular} &  \\ \hline
        Refine & \begin{tabular}[c]{@{}l@{}}%
        \texttt{Your first draft:} \\ 
        \texttt{```plaintext} \\ 
        \texttt{\{output\_complex\}} \\ 
        \texttt{```} \\ 
        \texttt{Errors:}\\ 
        \texttt{Your first transcription you made in ```plaintext block contains some errors.}\\ 
        \texttt{Task:} \\ 
        \texttt{Refine your first trasncription in ```plaintext block.}\\ 
        \texttt{Make sure to read the names of the people and the location as well as the dates}\\ 
        \texttt{and the numbers correctly.}\\ 
        \texttt{Transcribe as you see in the image.}\\ 
        \texttt{```plaintext}%
        \end{tabular} &  \\ \hline \hline
        Anti-error & \texttt{Even if it is hard to read the texts from the image, return as much as you can.} \\ 
        prompt & \texttt{You must read something. Do not return an apologetic message.} &  \\ \hline
        Post-  & \texttt{This is an output from you. Clean it such that we have no separators} \\ 
        processing & \texttt{and no comment from you:\{prompt\_parameter\}} &  \\ \hline
    \end{tabular}%
    }
    \label{tab:llm_prompts}
\end{table}

\begin{figure}[!ht]
    \centering
    \includegraphics[width=\linewidth]{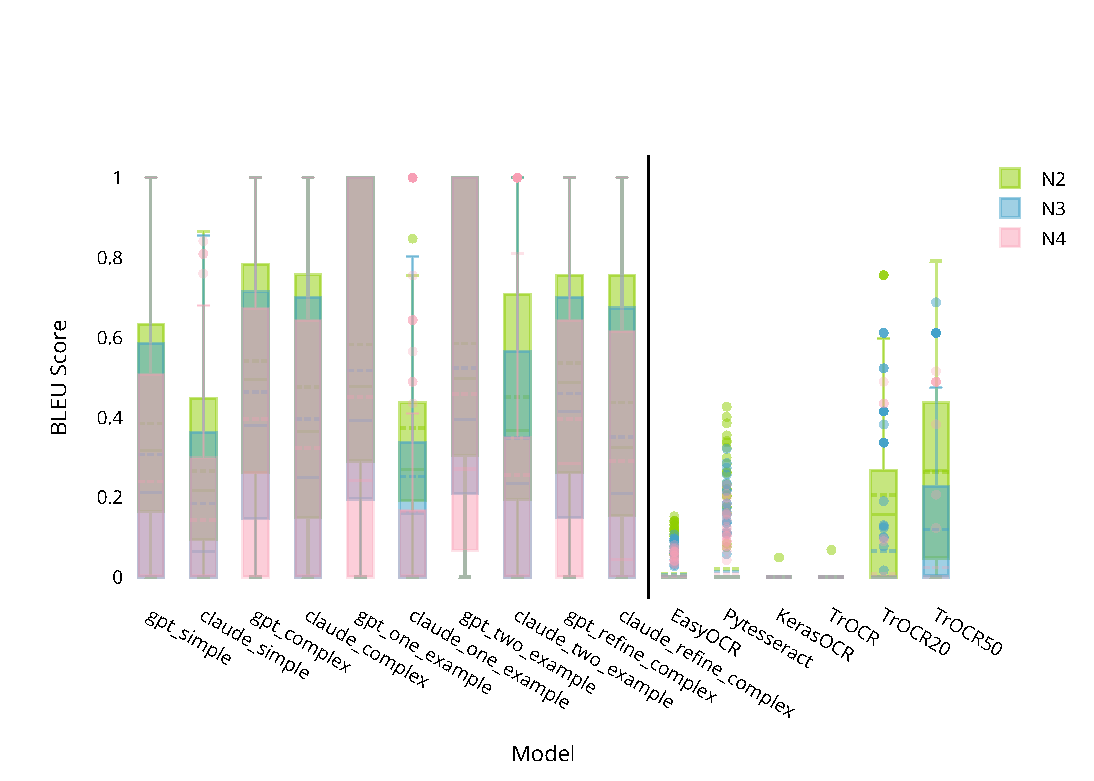}
    \caption{Comparisons between different maximum orders in n-grams: \raisebox{0.5ex}{\fcolorbox{lowered_green}{lowered_green!50}{\rule{0pt}{1pt}\rule{1pt}{0pt}}} bi-gram, \raisebox{0.5ex}{\fcolorbox{unidecoded_blue}{unidecoded_blue!30}{\rule{0pt}{1pt}\rule{1pt}{0pt}}} tri-gram and \raisebox{0.5ex}{\fcolorbox{normalized_pink}{normalized_pink!30}{\rule{0pt}{1pt}\rule{1pt}{0pt}}} 4-gram with line-by-line experiments. It illustrates that the lower the \textit{n} is, the higher the BLEU score is. Finding the proper maximum order is important since if one of the \textit{n}'s returns zero n-grams matched for a text, their geometric mean becomes zero. In that case, an output that managed to transcribe some words cannot be distinguished from an output that failed to transcribe completely.}
    \label{fig:ngrams_perline}
\end{figure}


\begin{figure}[!ht]
    \centering
    \vspace{-2em}
\begin{subfigure}[b]{0.9\linewidth}
    \centering
    \includegraphics[width=\linewidth]{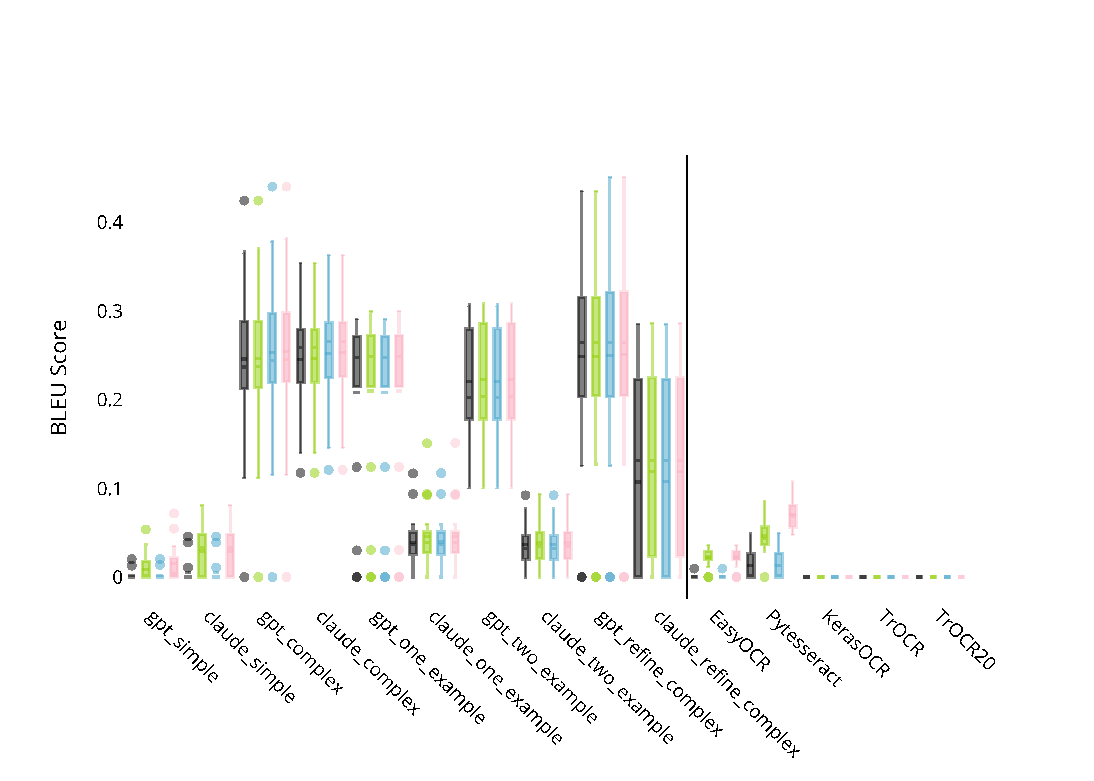}
    \caption{BLEU scores with the maximum order in the n-grams of 4.}
    \label{fig:bleu_whole_preprocessing}
\end{subfigure}
\begin{subfigure}[b]{0.9\linewidth}
    \centering
    \includegraphics[width=\linewidth]{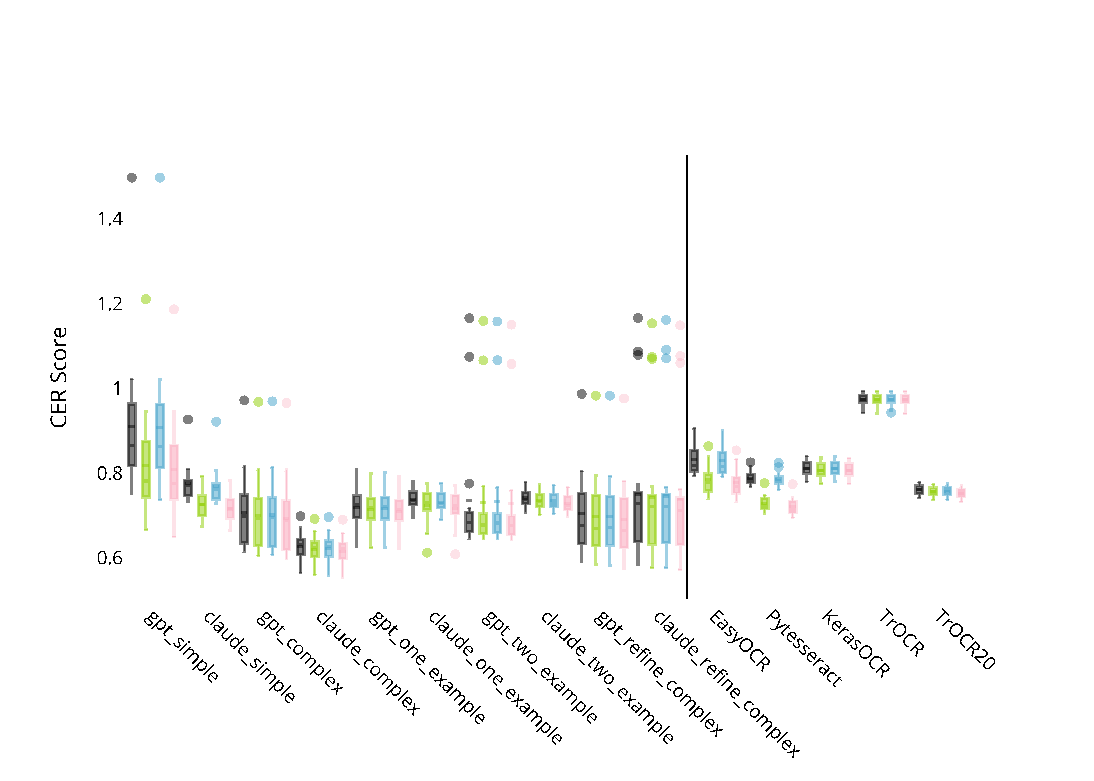}
    \caption{CER scores}
    \label{fig:cer_whole_preprocessing} 
\end{subfigure}
\caption{BLEU and CER score comparisons for each method on the whole-scan dataset: \raisebox{0.5ex}{\fcolorbox{onlystripped_gray}{onlystripped_gray!50}{\rule{0pt}{1pt}\rule{1pt}{0pt}}} Only stripped, \raisebox{0.5ex}{\fcolorbox{lowered_green}{lowered_green!50}{\rule{0pt}{1pt}\rule{1pt}{0pt}}} Lowered, \raisebox{0.5ex}{\fcolorbox{unidecoded_blue}{unidecoded_blue!30}{\rule{0pt}{1pt}\rule{1pt}{0pt}}} Unidecoded, and \raisebox{0.5ex}{\fcolorbox{normalized_pink}{normalized_pink!30}{\rule{0pt}{1pt}\rule{1pt}{0pt}}} Stripped, lowered and unidecoded. For each method, the mean score is in a dashed line, and the median is in a bold line.}
\label{fig:preprocessing_whole}
\end{figure}


\end{document}